\documentclass[lettersize,journal]{IEEEtran}
\ifCLASSOPTIONcompsoc
  \usepackage[nocompress]{cite}
\else
  \usepackage{cite}
\fi

\ifCLASSINFOpdf
  \usepackage[pdftex]{graphicx}
\else
\fi

\usepackage{amsmath}
\usepackage{bookmark}
\usepackage{subfig}
\DeclareMathAlphabet{\mymathbb}{U}{BOONDOX-ds}{m}{n}
{}
{}
\usepackage[linesnumbered,ruled]{algorithm2e}
\usepackage{threeparttable}
\usepackage{amsmath}
\usepackage{algpseudocode}
\newtheorem{Tlemma}{Lemma}

\newtheorem{Tdef}{Definition}

\usepackage{multirow}
\usepackage{xcolor}
\usepackage{subfig}

\SetCommentSty{mycommfont}    
\usepackage{pifont}
\let\oldding\ding
\renewcommand{\ding}[2][1]{\scalebox{#1}{\oldding{#2}}}

\begin{document}

\title{Scenario-Aware Control of Segmented Ladder Bus: Design and FPGA Implementation}

\author{Phu Khanh Huynh, Francky Catthoor, and Anup~Das
\IEEEcompsocitemizethanks{\IEEEcompsocthanksitem P. K. Huynh and A. Das are with the Department of Electrical and Computer Engineering, Drexel University, Philadelphia, PA, 19104.
E-mail: \{ph434, anup.das\}@drexel.edu. 
F. Catthoor is with National Technical University of Athens, Greece. Email: catthoor@microlab.ntua.gr.\protect\\
}
\thanks{Manuscript received Month DD, Year; revised Month DD, Year.}}



\maketitle

\begin{abstract}
Large-scale neuromorphic architectures consist of computing tiles that communicate spikes using a shared interconnect.
The communication patterns in these systems are inherently sparse, asynchronous, and localized, as neural activity is characterized by temporal sparsity with occasional bursts of high traffic.
These characteristics require optimized interconnects to handle high-activity bursts while consuming minimal power during idle periods. 
Among the proposed interconnect solutions, the dynamic segmented bus has gained attention due to its structural simplicity, scalability, and energy efficiency.
Since the benefits of a dynamic segmented bus stem from its simplicity, it is essential to develop a streamlined control plane that can scale efficiently with the network. 
In this paper, we present a design methodology for a scenario-aware control plane tailored to a segmented ladder bus, with the aim of minimizing control overhead and optimizing energy and area utilization.
We evaluated our approach using a combination of FPGA implementation and software simulation to assess scalability.
The results demonstrated that our design process effectively reduces the control plane's area footprint compared to the data plane while maintaining scalability with network size.

\end{abstract}


\section{Introduction}\label{sec:introduction}
Neuromorphic computing systems~\cite{kudithipudi2025neuromorphic} aim to replicate the computational principles of the brain by implementing spiking neural networks (SNNs) in hardware. 
These systems process information through discrete spike events rather than continuous signal transmission, resulting in a fundamentally different communication pattern compared to traditional computing architectures. Specifically, neuromorphic communication is sparse, asynchronous, and localized~\cite{yao2024spike}, where neurons generate spikes only when their membrane potential crosses a threshold. 
This event-driven behavior leads to temporal sparsity, meaning that most of the network remains idle for extended periods, with occasional bursts of activity when multiple neurons spike simultaneously. 
Consequently, the interconnect infrastructure in neuromorphic architectures must be optimized to accommodate this unique communication pattern, ensuring efficient data transfer during bursts while consuming minimal power during idle phases.

To facilitate communication in large-scale neuromorphic systems, many-core architectures are often employed, where processing elements (PEs) are interconnected using shared communication fabrics such as buses and networks-on-chip (NoCs)~\cite{truenorth, davies2018loihi}. 
While NoCs provide structured and scalable communication, they introduce significant overhead due to packet buffering, routing table management, and static link activity~\cite{effiong2022combined, xu2023improving}. 
These factors make NoCs less ideal for neuromorphic workloads, where traffic is sporadic. 
Instead, segmented bus architectures~\cite{balaji2019exploration} have been proposed as a more efficient alternative due to their ability to dynamically allocate communication pathways based on demand. 
By dividing the bus into segments and activating only the necessary paths for spike transmission, segmented buses can significantly reduce energy consumption while maintaining low latency.

Among various segmented bus implementations~\cite{balaji2022NeuSB}, the segmented ladder bus~\cite{huynh2024adiona} has been shown to offer the best energy efficiency and area utilization for runtime-configurable spiking traffic.
This dynamic segmentation balances adaptability, energy efficiency, and scalability, making it well-suited for neuromorphic applications.
However, its effectiveness depends on an efficient control mechanism.
Because the routing overhead is pushed fully to the control plane, a poorly designed controller can cause unnecessary state transitions, increasing area and power consumption and undermining the benefits of using a dynamic segmented bus.
To address this challenge, we are the first to propose and design a system scenario-aware control plane specifically for the segmented ladder bus architecture (to the best of our knowledge). 
This control plane is tailored to minimize control overhead and improve energy efficiency~\cite{catthoor2020system}.
By analyzing the traffic pattern of the application, our control plane can be adapted to minimize the frequently occurring communication scenarios. 
Essentially, by minimizing the number of active control scenarios, our design reduces the control related memory storage size and hence also reduces area footprint of the control plane to a negligible amount of the data plane. This makes it fully suitable for large-scale neuromorphic deployments.

To validate our approach, we implement the essential parts of the proposed control plane on FPGA and perform a combination of hardware-based and software-based simulations. 
We evaluate the impact of our scenario-aware optimization on area utilization and scalability. 
Experimental results demonstrate that our control plane design achieves significant reductions in resource overhead while maintaining reliable spike transmission. 

The rest of this paper is structured as follows: Section~\ref{sec:background} provides an overview of segmented ladder bus architecture and the existing deployment process for SNNs on such interconnect hardware.
Section~\ref{sec:solution} introduces our scenario-aware control methodology and optimization techniques. 
Section~\ref{sec:results} presents experimental results and analysis. 
Finally, the paper is concluded in Section~\ref{sec:conclusions}.

\section{Background}\label{sec:background}
This section presents the requisite background to comprehend the segmented ladder bus~\cite{huynh2024adiona}.
Furthermore, we review related works on the mapping of SNNs to neuromorphic hardware architectures, emphasizing the associated challenges and recent innovations in this domain.
\subsection{Segmented Ladder Bus}\label{sec:ladder_bus_basic}
The architecture of a segmented ladder bus is designed with the following characteristics:
\begin{itemize}
    \item The tiles are logically divided into two parallel rows, each comprising an equal number of tiles.
    \item A fixed number of parallel segmented bus lanes are placed between these tile rows.
    \item The tiles and parallel bus lanes are connected using criss-cross three-way segmented switches.
\end{itemize}

\begin{figure}[h!]
    \vspace{-5pt}
    \centering
    \centerline{\includegraphics[width=1\columnwidth]{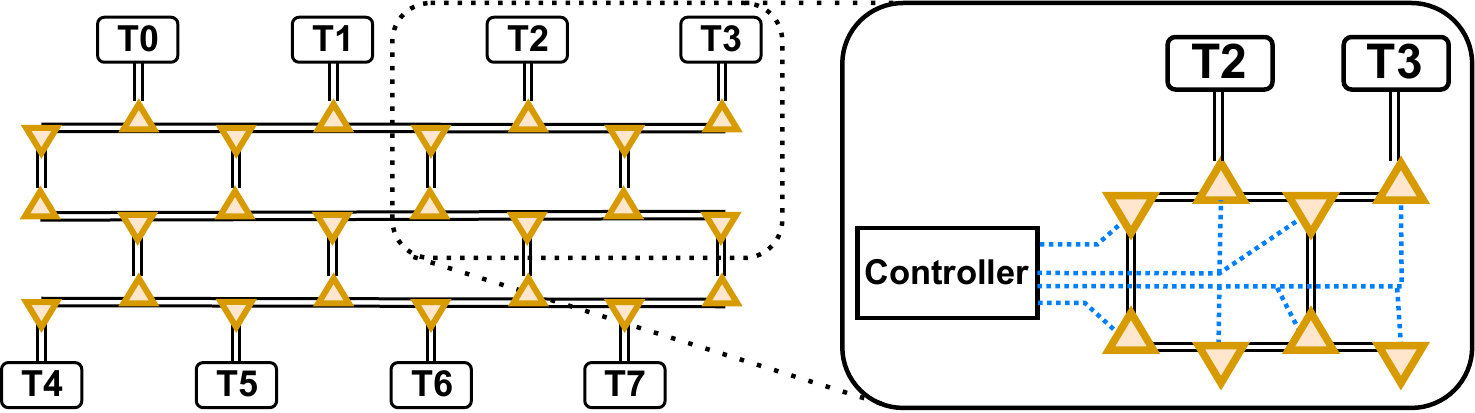}}
    \caption{Segmented ladder bus example with a switch controller.}
    \label{fig:ladder_bus_example}
    \vspace{-5pt}
\end{figure}

Figure~\ref{fig:ladder_bus_example} shows a segmented ladder bus with 8 tiles and 3 bus lanes, which can be scaled to meet application needs. This design offers several advantages, including high flexibility that enables communication between any pair of tiles, improved routing and fault-tolerance through multiple communication paths, run-time configurability allowing dynamic switch reconfiguration under congestion, and enhanced energy and area efficiency due to its bufferless architecture.

A segmented ladder bus uses a compile-time software framework alongside coarse-grain runtime hardware controllers for dynamic switch configuration. These distributed local controllers manage predefined switch scenarios and transmit control signals within their regions. In large-scale networks, many such controllers coordinate switch operation across the system. A key challenge of this dynamic approach is that it shifts all routing overhead to the control plane, making its design crucial for overall efficiency.


\subsection{Mapping and Scheduling SNNs on Segmented Ladder Bus}\label{sec:related_works}
\begin{figure}[h!]
    \vspace{-0pt}
	\centering
	\centerline{\includegraphics[width=1\columnwidth]{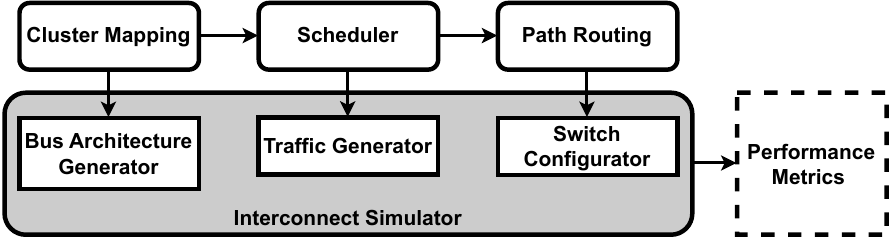}}
	\caption{Mapping and scheduling design flow.}
	\label{fig:design_flow}
        \vspace{-10pt}
\end{figure}

In our previous work~\cite{mustafazade2024clustering}, we have described the SNN application partitioning process in more details. 
Once an SNN application has been partitioned into clusters, in order to deploy it to a segmented ladder bus, we need to: map the clusters to actual hardware tiles; schedule spike traffic; and route the scheduled traffic. 
Each of these steps can be optimized to improve the hardware utilization of the network, guarantee performance of the application, and reduce energy consumption~\cite{huynh2025mapping}.
Figure~\ref{fig:design_flow} shows this process which is essential for running applications on segmented ladder bus.

This three step process can be summarized as follows:
\begin{itemize}
    \item \textbf{Mapping:} The cluster mapping problem in neuromorphic systems involves assigning neuron clusters to hardware tiles. Given the constraints of a segmented ladder bus, the mapping must minimize dynamic energy consumption and reduce intersecting communication paths to prevent delays or packet loss.
    \item \textbf{Scheduling:} Required when simultaneous connections exceed interconnect capacity, scheduling spreads traffic over time to prevent congestion while minimizing additional delays.
    \item \textbf{Routing:} Optimal routing minimizes energy and delay by guiding data through efficient paths. In bufferless segmented switches, it also prevents path crossings that cause collisions and spike loss. Additionally, routing balances traffic for better hardware utilization and provides fault tolerance through alternate paths.
\end{itemize}

\section{Proposed Control Plane Solution}\label{sec:solution}

The control plane of a segmented ladder bus must coordinate dynamic switch configurations in real time based on predefined communication patterns. Without optimization, the number of switching scenarios can grow rapidly with network size, increasing memory and hardware overhead. This motivates a scenario-aware control methodology to manage control information efficiently. The next section presents our proposed control plane design and algorithms, which minimize switching scenarios while ensuring correct communication.

\subsection{Local Controller Hardware Design}\label{sec:controller_hw_design}
The hardware implementation of a distributed local controller for segmented bus is designed to be simple and efficient. 
It operates using a local memory bank that stores predefined system scenarios tailored to the specific configuration of the bus segments it manages. 
During run-time, the controller dynamically selects the appropriate scenario based on the current communication demands. 
This scenario-based mechanism, identified at design-time and deployed at runtime, supports dynamic operation with a limited number of scenarios.
This mechanism works similarly to a distributed loop nest counter~\cite{raghavan2008distributed}. 
The central software control framework loads the entire loop nest of control scenarios into memories of local controllers. 
The scenarios are embedded as instructions for both regular (e.g., predictable switch sequences in static traffic patterns)) and irregular (e.g., conditional switch reconfiguration triggered by runtime events or spike activity) loop structures.
After loading, each controller autonomously steps through local control scenarios using a loop counter.
Once a scenario is chosen, the controller generates and transmits the corresponding control signals to the relevant switches.
By issuing the control signals from the controllers with the correct timing, the switches can be used to create paths that support multiple simultaneous connections inside the network, matching application requirements.
\begin{figure}[h!]
    \vspace{-5pt}
	\centering
	\centerline{\includegraphics[width=0.8\columnwidth]{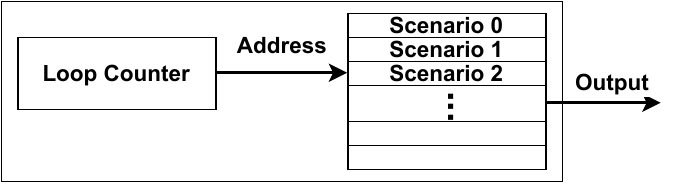}}
	\caption{Controller hardware design.}
	\label{fig:controller_design}
    \vspace{-10pt}
\end{figure}

\subsection{Control Plane Scenario Design}\label{sec:control_plane_scenario_design}
Using the cluster mapping data, we extract the required communication paths and group them into non-intersecting scenarios to minimize the total number of switching cases. We first apply a greedy algorithm (Algorithm~\ref{alg:greedy_grouping}) that iteratively assigns each path to the first group where it does not intersect with others, forming a new group if needed. This continues until all paths are assigned, with each group forming a distinct scenario for the control plane. While this ensures conflict-free grouping, it is non-optimal and still produces a relatively high number of scenarios, as shown in Section~\ref{sec:results}.
\vspace{-1em}
\begin{algorithm}[h!]
    \scriptsize{
    \KwIn{Cluster mapping data with communication paths}
    \KwOut{Set of non-conflicting switching scenarios}

    Initialize scenario set $S \gets \emptyset$\;
    \ForEach{path $p$ in the path list}{
        placed $\gets$ \textbf{false}\;
        \ForEach{scenario $s \in S$}{
            \If{no path in $s$ intersects with $p$}{
                Add $p$ to $s$\;
                placed $\gets$ \textbf{true}\;
                \textbf{break}\;
            }
        }
        \If{not placed}{
            Create a new scenario $s_{\text{new}} \gets \{p\}$\;
            Add $s_{\text{new}}$ to $S$\;
        }
    }
    \textbf{return} $S$\;
    }
    \caption{Greedy Grouping}
    \label{alg:greedy_grouping}
\end{algorithm}
\vspace{-1em}
\vspace{-1em}
\begin{algorithm}[h!]
    \scriptsize{
    \KwIn{Cluster mapping data with communication paths}
    \KwOut{Set of non-conflicting switching scenarios}
    
    Construct graph $G(V, E)$ where vertices $V$ represent communication paths\;
    \ForEach{$(p_i, p_j) \in V \times V$}{
        \If{$p_i$ and $p_j$ intersect}{
            Add edge $(p_i, p_j)$ to $E$\;
        }
    }
    Initialize scenario set $S = \emptyset$\;
    \While{$V \neq \emptyset$}{
        Find the maximal clique $C \subseteq V$\;
        \ForEach{$v \in C$}{
            Assign $v$ to a distinct scenario in $S$\;
        }
        Remove $C$ from $G$\;
        \If{any remaining vertices cannot be assigned to existing scenarios}{
            Create a new scenario\;
        }
    }
    \textbf{return} $S$\;
    }
    \caption{Max Clique Grouping}
    \label{alg:clique_grouping}
\end{algorithm}
\vspace{-1em}
Alternatively, we design a more exploratory algorithm that reduces greediness and takes into account a broader search space.
It works as follows: Communication paths can be represented as vertices in a graph, where an edge is established between any two vertices if their corresponding paths intersect. 
The lower bound for the number of required scenarios is determined by the maximal clique of this conflict graph, as all paths within this clique mutually intersect and must therefore be assigned to separate scenarios.
To achieve this, a computationally efficient algorithm is employed to identify the maximal clique. 
The vertices within this clique are then distributed into distinct scenarios. 
Subsequently, the identified clique is removed from the graph, and the process iterates: identifying the next maximal clique, partitioning its vertices into scenarios while ensuring no path intersections within a single scenario, and repeating the cycle.
If, at any stage, the remaining vertices cannot be accommodated within the existing scenarios, a new scenario is created. This iterative process continues until all vertices have been removed.

\textbf{Time Complexity Analysis of Algorithm~\ref{alg:clique_grouping}:}
Let $n$ be the number of communication paths and $m$ the number of edges in the intersection graph $G(V, E)$. Constructing $G$ takes $O(n^2)$ time by checking all path pairs. The main loop runs up to $O(n)$ times, and each iteration performs maximal clique extraction—an NP-hard task with worst-case time $O(3^{n/3})$ using exact algorithms like Bron–Kerbosch~\cite{johnston1976cliques}, with graph updates $O(n + m)$ being negligible in comparison. The overall complexity is $O(n^2 + n \cdot 3^{n/3})$.


\section{Results and Discussions}\label{sec:results}
Table~\ref{tab:apps} reports the machine learning applications utilized in the assessment of designing scenario-aware control plane for segmented ladder bus.
For each model, we provide detailed information on the total number of clusters, the average connection degree, network density (the connections between clusters are directional), and the largest connection degree.
Six realistic and four synthetic applications were used. 
\begin{table}[!h]
\renewcommand{\arraystretch}{1.0}
\centering
\scriptsize
\resizebox{\columnwidth}{!}{%
\begin{tabular}{c|c|c|c|c}
\hline
\textbf{Application} & \textbf{\# Clusters} & \textbf{Avg. Degree} & \textbf{Network Density} & \textbf{Largest Degree} \\
\hline
    mnist & 11 & 1.64 & 0.135 & 6 \\
    LeNet~\cite{lecun2002gradient} & 14 & 2.93 & 0.225 & 11 \\
    fashion-mnist~\cite{xiao2017fashion} & 24 & 5.33 & 0.230 & 8 \\
    cifar10 & 26 & 5.44 & 0.210 & 8 \\
    emnist~\cite{cohen2017emnist} & 30 & 5.37 & 0.185 & 8 \\
    synth\_40 (160) & 40 & 4.00 & 0.105 & 7 \\
    synth\_40 (292) & 40 & 7.30 & 0.185 & 14 \\
    synth\_60 (348) & 60 & 5.80 & 0.100 & 12 \\
    synth\_60 (772) & 60 & 12.87 & 0.220 & 21 \\
    ResNet~\cite{he2016deep} & 96 & 11.13 & 0.115 & 78 \\
\hline
\end{tabular}%
}
\caption{Applications used to evaluate control plane scaling.}
\label{tab:apps}
\vspace{-5pt}
\end{table}

\textbf{FPGA implementation:} To estimate control plane resource usage, we implemented the small and medium applications on FPGA with their respective number of clusters.
The segmented ladder bus was configured with a number of lanes equal to the square root of the tile count, each 32 bits wide.
Scenario counts were derived using Algorithm~\ref{alg:clique_grouping}.
The results are presented in Table~\ref{tab:plane_utilization}.
From this table, we can see clearly that the control plane utilizes much fewer resources than the data plane. 
In all applications, the control plane utilize less than 10\% of the total resources, with an average utilization of 6.5\%.
These practical results show that, in small to medium-scale applications, the control overhead remains minimal compared to the data plane.

\begin{table}[!h]
\renewcommand{\arraystretch}{1.0}
\centering
\footnotesize 
\resizebox{\columnwidth}{!}{%
\begin{tabular}{c|c|c|c|c|c}
\hline
\textbf{App} & \textbf{\# Lanes} & \textbf{\# Scen.} & \textbf{D-Plane CLBs} & \textbf{C-Plane CLBs} & \textbf{C-Plane Util.} \\
\hline
mnist & 3 & 8 & 3277 & 73 & 2.18\% \\
LeNet & 4 & 13 & 3503 & 204 & 5.50\% \\
fashion mnist & 5 & 24 & 5722 & 543 & 8.67\% \\
cifar10 & 5 & 23 & 6416 & 548 & 7.87\% \\
emnist & 5 & 26 & 7247 & 645 & 8.17\% \\
\hline
\end{tabular}%
}
\caption{Data and control plane utilizations.}
\label{tab:plane_utilization}
\vspace{-10pt}
\end{table}

\begin{figure}[h!]
        \vspace{-5pt}
	\centering
	\centerline{\includegraphics[width=1\columnwidth]{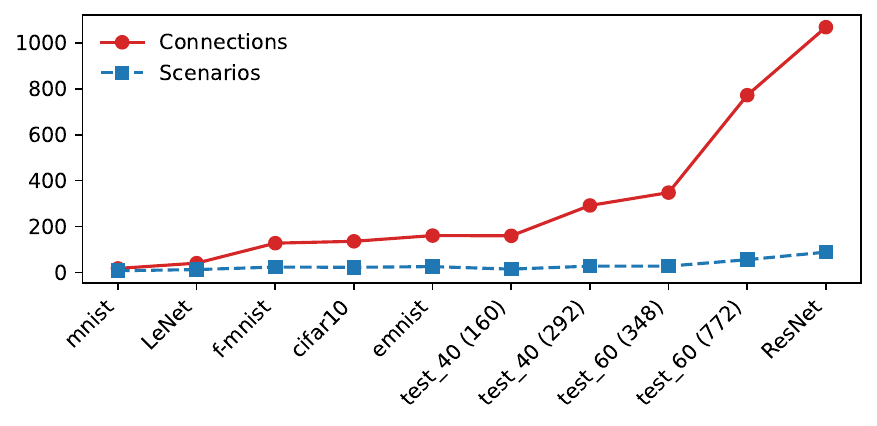}}
	\caption{Number of connections and scenarios scaling.}
	\label{fig:scene_scale}
        \vspace{-5pt}
\end{figure}

\textbf{Simulation scalability analysis:} For larger networks that exceed available FPGA resources, we perform an analytical evaluation of the control plane. Specifically, we examine how the number of scenarios—a key scaling parameter—grows relative to the data plane.
Figure~\ref{fig:scene_scale} compares the total number of connections with the number of scenarios produced by Algorithm~\ref{alg:clique_grouping}. 
The results show that scenario count grows much more slowly than both the number of connections and clusters. 
In contrast, the data plane scales proportionally with the number of clusters and the number of bus lanes. 
This indicates that as the network size increases, the control plane area remains within a manageable range and grows more slowly than the data plane, making it negligible in area and energy overhead.
As the network scales, each scenario encodes increasingly sparse information with many zero values, creating opportunities for compression that further reduce control memory size.
Moreover, scenarios are distributed across local controllers, each handling fewer switches and storing smaller scenario sets, which enhances control plane scalability.

\begin{figure}[h!]
        \vspace{-5pt}
	\centering
	\centerline{\includegraphics[width=1\columnwidth]{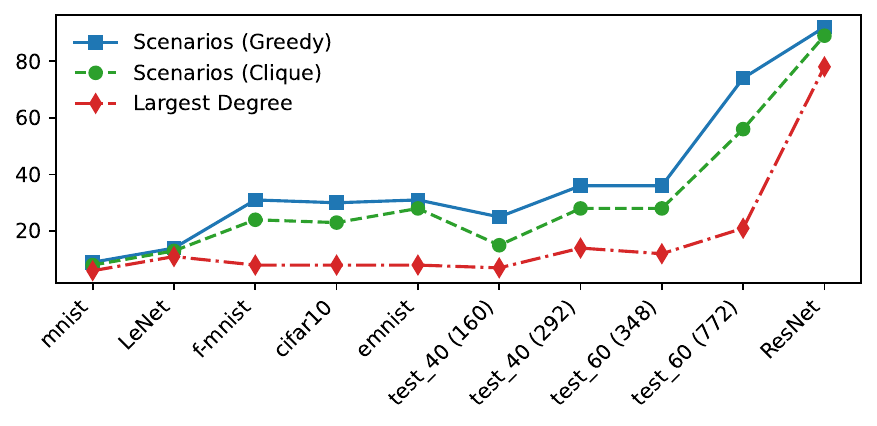}}
	\caption{Scenario scaling for different algorithms.}
	\label{fig:algo_scene}
        \vspace{-5pt}
\end{figure}

We implemented both the greedy and maximum clique scenario grouping algorithms, with results shown in Figure~\ref{fig:algo_scene}. The maximum clique approach consistently yields fewer scenarios than the greedy method. The largest node degree provides a theoretical lower bound, as unicast connections with shared sources or destinations must be placed in separate groups. Notably, the maximum clique algorithm approaches this theoretical limit, particularly in low-density networks.

\section{Conclusions}\label{sec:conclusions}
This work proposed a scenario-aware control methodology for the segmented ladder bus, a dynamic interconnect designed for large-scale neuromorphic systems. 
By leveraging compile-time traffic analysis, our control plane design minimizes the number of switching scenarios, reducing control memory and hardware overhead. 
Our FPGA implementation and simulation analysis show that the control plane remains lightweight, using under 10\% of the total network resources. Scalability analysis further indicates that control complexity grows more slowly than network connectivity. 
Overall, our scenario-aware control plane design offers an efficient and scalable solution for neuromorphic interconnects.

\ifCLASSOPTIONcompsoc
  \section*{Acknowledgments}
\else
  \section*{Acknowledgment}
\fi

This work is supported by US DOE Award DE-SC0022014 and the US NSF Award CCF-1942697.

\bibliographystyle{IEEEtran}
\bibliography{external,disco}


\end{document}